\documentclass[10pt,twocolumn,letterpaper]{article}

\usepackage{cvpr}
\usepackage{times}
\usepackage{epsfig}
\usepackage{graphicx}
\usepackage{amsmath}
\usepackage{amssymb}

\usepackage[breaklinks=true,bookmarks=false]{hyperref}

\cvprfinalcopy 

\def\cvprPaperID{****} 

\begin{document}

\title{Transcription-Enriched Joint Embeddings\\for Spoken Descriptions of Images and Videos}

\author{Benet Oriol, Jordi Luque and Ferran Diego\\
Telefonica Research\\
Barcelona, Catalonia / Spain\\
{\tt\small jordi.luqueserrano@telefonica.com}
\and
Xavier Giro-i-Nieto\\
Universitat Politecnica de Catalunya (UPC)\\
Barcelona, Catalonia / Spain\\
{\tt\small xavier.giro@upc.edu}
}

\maketitle
\begin{abstract}
In this work, we propose an effective approach for training unique embedding representations by combining three simultaneous modalities: image and spoken and textual narratives. The proposed methodology departs from a baseline system that spawns a embedding space trained with only spoken narratives and image cues. Our experiments on the EPIC-Kitchen and Places Audio Caption datasets show that introducing the human-generated textual transcriptions of the spoken narratives helps to the training procedure yielding to get better embedding representations. The triad speech, image and words allows for a better estimate of the point embedding and show an improving of the performance within tasks like image and speech retrieval, 
even when text third modality, text, is not present in the task.
\end{abstract}


\section{Introduction}
Deep neural networks have become increasingly popular in very different contexts, tasks and
modalities, such as vision, audio and language \cite{rulethemall}. 
Multimodal joint embeddings are unified representations for different media types
which are generated by modality-specific neural encoders.
The parameters of these encodes are often learned with a triplet loss (or similar) that aligns multimodal representations from the same data into the same region of the feature space.
For instance, an image of a cat would be mapped by an image encoder to a similar embedding to the one generate by a language encoder of the word "cat". 

This work explores the gain obtained when adding the textual transcriptions to the joint embeddings learned from egocentric activity images and their spoken narratives.
This addition of a the textual modality helps into obtaining richer and more robust representations, which we test in an bi-directional audio from/to video frame  retrieval task. This gain is obtained at the cost of collecting the textual transcriptions of the spoken narratives in the training set. Notice though that this transcriptions are not required at test time, as the features are obtained from whether the speech or visual data, only.

\section{Previous work}
Multimodal features  can be of many different kinds. One of the most simple approaches can be concatenating the representations of all modalities into a single multimodal vector. However, this work focuses on \textit{similarity models} that combine different neural encoders that independently map each modality into a single representation space.

DeViSE \cite{prev1} was one of the first works of similarity models. Their approach consisted of a simple linear transforms that maps from image and text space to a common embedding space, where the inner product or cosine distance between corresponding concepts in different modalities is minimized. Later, Frome et al. \cite{prev2} proposed more complex mappings to these shared embedding spaces. 
Pan et al. \cite{prev3} used a similar approach, but on videos, and using a recurrent neural network. Suris et al. \cite{prev4} also worked in videos domain with features pooled for each video clip.
Aytar et al. \cite{aytar2017see} proposed a triple modality embedding, using two pair losses, one for the image and text branch and one for the image and audio branch. Their sound branch represent ambient sound, and they use retrieval metrics to prove the improvement of their system with respect to image-sound-text alignement models. 
Harwath, Torralba et al. \cite{harwath2016unsupervised} used fully convolutional models to map audio descriptions and images into a shared embedding space. 
Similar to that last one, Harwath, Recasens et al. \cite{jointly} also mapped speech narrations of images and audio descriptions to a common embedding space, with the main difference that the image and audio feature models do not pool the localized into a single vector, but outputs features that keep the spatial and temporal coordinates. 

\section{Methodology}
This work builds upon the neural architecture proposed by Harwath, Recasens et al. \cite{jointly}. 
This architecture is fully convolutional for both the image and speech encoders, which 
allows localizing the correlations between images and spoken narratives in both space and time . 
The baseline loss function used by \cite{jointly} is 
\begin{equation} \label{baseline_loss}
\begin{split}
L = \sum_{i=1}^{B}(&\textup{max}(0, S_{IA}(I_i, A_j) - S_{IA}(I_i, A_i) +\eta) + \\
 &\textup{max}(0, S_{IA}(I_k, A_i) - S_{IA}(I_i, A_i) + \eta )),
\end{split}
\end{equation}
being $B$ the size of the minibatch, $I_i$ and $A_i$ the image and audio features of the $i$th element of the batch, $j, k \neq i$ the impostor indexes, $\eta$ the margin parameter, and $S_{IA}(I, A)$ a similarity function between image and audio features. This similiarity function is further discussed in Section \ref{sec:similiarity}.

This training loss is a ranking-based criterion that intuitively aims at maximizing the similarity of corresponding image-audio pairs, and minimizing the similarity of non-corresponding ones. For our approach, and in order to incorporate textual embeddings in the training, we extend the previous equation, that involves image and audio, to three similar instances of the same loss. Each of these three ranking losses corresponds to a modality combination: image and audio, image and text or text and audio. 
The three loss functions are combined as

\begin{equation} \label{complete_loss}
\begin{split}
L = \sum_{i=1}^{B}(&\textup{max}(0, S_{IA}(I_i, A_j) - S_{IA}(I_i, A_i) +\eta) + \\
 &\textup{max}(0, S_{IA}(I_k, A_i) - S_{IA}(I_i, A_i) + \eta ) \\
 &\textup{max}(0, S_{IT}(I_i, T_l) - S_{IT}(I_i, T_i) + \eta ) \\
 &\textup{max}(0, S_{IT}(I_m, T_i) - S_{IT}(I_i, T_i) + \eta ) \\
 &\textup{max}(0, S_{TA}(T_i, A_n) - S_{TA}(T_i, A_i) + \eta ) \\
 &\textup{max}(0, S_{TA}(T_o, A_i) - S_{TA}(T_i, A_i) + \eta )),
\end{split}
\end{equation}
being $l,m,n,o \neq i$ also impostor indices from inside the minibatch.

This loss function will maximize the similarity between corresponding image-audio pairs, audio-text pairs and image-text pairs and minimize the similiarity between non-corresponding ones. With it, we will be able to train three different branches, mapping corresponding signals from three different modalities into similar points in the embedding space.

\subsection{Image Encoder}
As proposed in \cite{jointly}, the image encoder follows a VGG-16 \cite{VGG} architecture, up to the \texttt{Conv-5} module, without the maxpool included. On top of that, we add a convolutional layer to map from the 512-dim space VGG output to the desired embedding size space.

\subsection{Speech Encoder}
As proposed in \cite{jointly}, speech is firstly transformed into log Mel filter bank spectro-grams, with 25ms Hamming window and 10ms time shift. 
After that, we use a fully convolutional architecture, to map to the embedding space. 
This model consists of 5 convolutional layers and max poolings.

\subsection{Text Encoder}
For the text model, we use an off-the-shelf pretrained BERT \cite{devlin2018bert} model, state-of-the-art in natural language feature extraction. We use the Huggingface implementation \footnote{\url{https://github.com/huggingface/transformers}} and the \texttt{bert-base-uncased} version.

\subsection{Similarity functions}
\label{sec:similiarity}
The three models output features that are localized in the space, time or sentence position. In other words, they do not squeeze, respectively, an image, an audio or sentence into single vector. The image model outputs a $N_r\times N_c\times EmbSize$ tensor, being $N_r\times N_c$ the spatial dimensions, and $EmbSize$ the dimensionality of the embedding space. The audio model outputs a $N_a\times EmbSize$ tensor, and the text model a $N_w\times EmbSize$ sized tensor, where $N_{w}$ and $N_{a}$ depend on the length of the text and audio. For this reason, computing a similarity between two feature maps is not as straight-forward as, for example, computing a cosine similarity between two vectors. 

In order to compute a similarity between them, we first compute a matchmap. The matchmap is the result of computing a cosine similarity between two feature maps in the embedding dimension. More formally, if $I[r, c]$  is the image feature vector at the image embedding width and height $r$ and $c$, and $A[t]$ is the audio feature vector at audio embedding time $t$, we can define the image-audio matchmap $M_{IA}[r, c, t]$ with equation \ref{eq:matchmap}. 

\begin{equation} \label{eq:matchmap}
    M_{IA}[r, c, t] = \langle I[r, c] , A[t]\rangle
\end{equation}
\begin{equation} \label{eq:matchmap_it}
    M_{IT}[r, c, w] = \langle I[r, c] , T[w]\rangle
\end{equation}
\begin{equation} \label{eq:matchmap, ta}
    M_{TA}[t, w] = \langle T[w] , A[t]\rangle
\end{equation}

being $\langle\cdot , \cdot\rangle$ the inner product operation. Note that while $I[r, c]$ and $A[t]$ are feature vectors (both with size equal to the embedding dimesionality we chose), each element of the matchmap is a scalar value, that expresses the similarity between the image embedding vector at $[r, c]$ and the audio embedding at $[t]$. Similarly to the image-audio matchmap, we will define the text-audio matchmap and image-text matchmap with Equations \ref{eq:matchmap, ta} and \ref{eq:matchmap_it}, respectively.

The matchmap can be useful to co-localize concepts in space, time and text, but in order to train the models with our loss, we need to express similarity as a scalar value, getting an overall similarity out of an image-audio matchmap. Harwath, Recasens et al. \cite{jointly} proposed three different similarity functions for this purpose. This work explore two of them: SIMA (Summation over Image and Maximum over Audio) and MISA (Maximum over Image and Summation over Audio), following Equations \ref{eq:sima} and \ref{eq:misa}:

\begin{equation}\label{eq:sima}
    SIMA (M_{IA}) = \frac{1}{N_cN_r}\sum_{r=0}^{N_r-1}\sum_{r=0}^{N_r-1}\max_{t}M_{IA}[r, c, t]
\end{equation}
\begin{equation}\label{eq:misa}
    MISA (M_{IA}) = \frac{1}{N_a}\sum_{r=0}^{N_a-1}\max_{r, c}M_{IA}[r, c, t]
\end{equation}
Following this idea, we use SIMT and STMA for image-text matchmaps and text-audio matchmaps, represented in equations \ref{eq:simt} and \ref{eq:stma}, respectively.
\begin{equation}\label{eq:simt}
    SIMT (M_{IT}) = \frac{1}{N_cN_r}\sum_{r=0}^{N_r-1}\sum_{r=0}^{N_r-1}\max_{t}M_{IT}[r, c, w]
\end{equation}
\begin{equation}\label{eq:stma}
    STMA (M_{TA}) = \frac{1}{N_w}\sum_{r=0}^{N_w-1}\max_{t}M_{TA}[w, t]
\end{equation}

\section{Experiments}

The impact of adding the textual transcription of spoken narrations in the learned joint embeddings is assessed for a cross-modal retrieval task in two datasets. The details of the experiments and results obtained are presented in this section.

\subsection{Datasets}

\subsubsection{EPIC Kitchens}

The EPIC Kitchens dataset \cite{EPICkitchens} contains egocentric videos of kitchen procedures. Moreover, this dataset also includes the spoken narrations of the actions performed by the user wearing the camera, and clean transcriptions of those narrations in natural language (text).

In order to adapt this dataset for our task, we needed to complete some preprocessing steps. Firstly, as we mentioned, this dataset has actions and speech action narrations. However, they are annotated independently, with no explicit correspondence between both of them.
For this reason, we had to align the two sets, using heuristics on the natural language transcription, which is an available field both at the narration annotations and at the action narration annotations. The heuristics included a string comparison (verb stemming, checking string equality and checking inclusion of the narration string into the action string) and a simple alignment algorithm. 

With this action-narration correspondence, now we proceeded to segment each kitchen video into smaller clips and the spoken narration into smaller audio clips, using the corresponding timestamps in the action and action narrations annotations. This way, we ended up with tuples of video clips containing a single action and audio narrations containing this single action's description, thus creating corresponding video-audio pairs. Moreover, as we mentioned, each narration also had its clean natural language representation, therefore we actually obtained video-audio-text tuples.

The system works with static image, but now we have video clips. For each video clip we chose N frames, and generate N static image tuples with those. The way we chose those N frames was different in train and validation and test. In the train split, we selected N+2 equally-spaced frames across all video clip, and discarded the first and the last. We did it this way to get the maximum visual diversity among all N frames, but we discarded the first and last since we saw that it normally did not contain much useful information related to the action in the clip. We set N=5 for training. For the validation and test sets, we kept N=1 and select the middle frame in the clip, in order to maximize visual correspondence and keep a small validation set. We will see how validation computation cost grows with $O(2)$ with the validation set size.

With respect to the audio, we decided to adjust the timestamps to 0.3s before the annotated ones, since we observed the beginning of the narrations was being chopped when using the annotated timestamps. Moreover, we limited the length of the narrations to maximum 3s and minimum 0.1s. Also we only took the ones in English.


Before selecting the N frames out of each clip, this left us with a total of 15145 clip-narration pairs. We randomly split the clip-narration pairs into 14.000 training examples, 600 validation examples and 545 test examples. After that, we use the previously explained frame picking rules to obtain 70000 training tuples, 600 validation tuples and 545 test tuples. Again, when we talk about tuples, we refer to image-narration-text tuples.

\subsubsection{Places Audio Caption dataset}

The Places Audio Caption dataset \cite{harwath2016unsupervised} \cite{jointly} is a collection of approximately 400k audio caption descriptions obtained via Amazon Mechanical Turk. The described images are from the Places 205 image dataset. This dataset is much larger than EPIC kitchens and has a much bigger variety in concepts and complexity of narrations.

The train split consists of 402385 examples and validation split has 1000 examples.

\begin{table*}[!ht]
\begin{center}
\begin{tabular}{|l|l|c|c|c|c|c|c|}
\hline
Text & Dataset & Img R@1 & Img R@5 & Img R@10 & Audio R@1 & Audio R@5 & Audio R@10  \\
\hline\hline
No & EPIC Kitchens & 0.178 & 0.472 & 0.639 & 0.182 & 0.481 & 0.637\\
Yes & EPIC Kitchens & \textbf{0.193} & \textbf{0.514} & \textbf{0.686} & \textbf{0.183} & \textbf{0.494} & \textbf{0.677} \\
\hline
No & Places & 0.061 & 0.211 & 0.312 & 0.046 & 0.171 & 0.262 \\
No & Places * & 0.079 & 0.225 & 0.314 & 0.057 & 0.191 & 0.291 \\
Yes & Places & \textbf{0.116} & \textbf{0.278} & \textbf{0.394} & \textbf{0.072} & \textbf{0.24} & \textbf{0.338} \\
\hline
\end{tabular}
\end{center}
\caption{In this table we present the recall scores in both datasets and with or without adding text at train time. The difference between the 3rd and 4th line (marked with *) is that the 4th are the results published at \cite{jointly} while the 3th are our results of the same experiment. We used the implementation they provided, but couldn't have the same experimental setup, reducing the batch size from 128 to 80.
Results are on test dataset for EPIC and validation for Places}
\label{table:retrieval}
\end{table*}


\subsection{Training Details}
All models were trained with SGD, with learning rate of 0.001, decaying by a ratio of 10 every 70 epochs, and a momentum of 0.9. When not using text, models were trained with the loss function from equation \ref{baseline_loss} and when using text, models are trained with the loss function from equation \ref{complete_loss}. Regarding the similarity functions, we used SIMA, SIMT and STMA for EPIC kitchens and MISA, MIST and STMA for the Places dataset.
For EPIC kitchens, we used a batch size of 30 and for Places a batch size of 80. This because we used machines with more computational resources to run the experiments of the biggest dataset.
The BERT embedding was kept frozen, so the weights were not updated. The size of the embedding was set to 768, to match the BERT output dimensions and did not need to train anything on top of it.

\subsection{Cross-modal Retrieval}
The task used to assess the quality of the embeddings was image and audio retrieval. In other words, we computed the similarity between all audios and all images in the validation split, and perform speech narrations retrieval queried by image and retrieval of image queried by speech narrations, using the similarity as ranking score to retrieve images.


Table \ref{table:retrieval} shows that adding text at training time improves all recall scores. We think this is due to two main reasons. Firstly, textual transcriptions can be understood as a clean version on audio data, and it encodes in a much cleaner way the concept in the video clip. For this reason, it is no surprise that it helps as an intermediate modality between audio and image.
Secondly, pretrained BERT is the state-of-the-art in text embeddings. This model had already discovered concepts during its pretraining. For this reason, we do not need the audio and image branches (trained from scratch) to \textit{discover} the concepts on their own, and find BERT as almost a ground truth embedding they need to learn the mapping to.


\section{Conclusions}
This work shows how textual transcriptions can enrich the quality of the representations learned for joint embedding spaces between images and speech. The textual representations improve the image and speech encoders, as they help to train more robust 
image and speech
embeddings that improve the recall metrics in the tasks of speech narration retrieval queried by image, and viceversa.
We also point out that the egocentric nature of speech and vision, could help the retrieval task.

\section*{Acknowledgements}
This work was supported by the the Industrial Doctorate 2017-DI-011 from the Government of Catalonia, and the Spanish Ministry of Economy and Competitiveness through the European Regional Development Fund (TEC2016-75976-R). We gratefully acknowledge the support of NVIDIA Corporation with the donation of GPUs used in this work.

{\small
\bibliographystyle{ieee_fullname}
\bibliography{egbib_final}
}

\end{document}